\documentclass[letterpaper, 10 pt, conference]{ieeeconf}

\IEEEoverridecommandlockouts
\usepackage{bm}
\usepackage{booktabs}
\usepackage{graphicx} %
\usepackage{subcaption}
\usepackage{colortbl}
\usepackage{amsmath}
\usepackage{amsfonts}
\usepackage{amssymb}
\usepackage{textcomp}
\usepackage[colorlinks,allcolors=blue]{hyperref}
\usepackage{color}
\usepackage{svg}
\usepackage{tikz}

\newcommand{\Reals}{\mathbb{R}}
\newcommand{\inputdim}{D}
\newcommand{\numobs}{N}
\newcommand{\Normal}{\mathcal{N}}
\newcommand{\y}{\mathbf{y}}
\newcommand{\x}{\mathbf{x}}
\newcommand{\X}{X}

\renewcommand{\u}{\mathbf{u}}
\newcommand{\z}{\mathbf{z}}
\newcommand{\transpose}{\intercal}
\newcommand{\calcd}{\mathrm{d}}
\newcommand{\argmin}{\mathrm{argmin}}
\newcommand{\shortrightarrow}{%
\parbox{.18cm}{\tikz{\draw[->](0,0)--(.15cm,0);}}
}

\newcommand{\meanfunction}{\mu}
\newcommand{\kernelfn}{k}
\newcommand{\kernelhyps}{\theta}
\newcommand{\kernelmatrix}{K}
\newcommand{\kernelvector}{\mathbf{k}}

\newcommand{\f}{\mathbf{f}}
\newcommand{\numinducing}{M}
\newcommand{\Zall}{Z_{\mathrm{all}}}

\newcommand{\parent}{{\mathrm{par}}}

\newcommand{\dcomm}{d_{\mathrm{comm}}}

\newcommand{\factor}{\phi}
\newcommand{\nei}{\mathrm{nei}}
\newcommand{\vargeneric}{\x}
\newcommand{\numvars}{N}
\renewcommand{\message}{m}
\newcommand{\messagevartofac}[2]{\message_{\vargeneric_{#1}\shortrightarrow\factor_{#2}}}
\newcommand{\messagefactovar}[2]{\message_{\factor_{#1}\shortrightarrow\vargeneric_{#2}}}
\newcommand{\info}{\eta}
\newcommand{\Prec}{\Lambda}
\newcommand{\infomessage}[2]{\info_{#1\shortrightarrow#2}}
\newcommand{\Precmessage}[2]{\Prec_{#1\shortrightarrow#2}}

\title{\LARGE \bf
A Distributed Gaussian Process Model for Multi-Robot Mapping
}
\author{Seth Nabarro$^{1*}$, Mark van der Wilk$^{2}$ and Andrew J. Davison$^{1}$
\thanks{$^{1}$Dyson Robotics Lab, Imperial College London}%
\thanks{$^{2}$Department of Computer Science, University of Oxford}%
\thanks{$^*$ {\tt\small sdn09@ic.ac.uk}}}%
\begin{document}
\maketitle
\thispagestyle{empty}
\pagestyle{empty}
\begin{abstract}
    We propose \emph{DistGP}: a multi-robot learning method for collaborative learning of a global function using only local experience and computation. We utilise a sparse Gaussian process (GP) model with a factorisation that mirrors the multi-robot structure of the task, and admits distributed training via Gaussian belief propagation (GBP). Our loopy model outperforms Tree-Structured GPs \cite{bui2014tree} and can be trained online and in settings with dynamic connectivity. We show that such distributed, asynchronous training can reach the same performance as a centralised, batch-trained model, albeit with slower convergence. Last, we compare to DiNNO \cite{yu2022dinno}, a distributed neural network (NN) optimiser, and find DistGP achieves superior accuracy, is more robust to sparse communication and is better able to learn continually.
\end{abstract}
\section{Introduction}
\begin{figure*}[t!]
    \centering
    \includegraphics[width=0.9\textwidth]{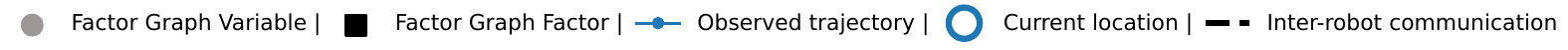}
     \begin{subfigure}[t]{0.43\textwidth}
        \centering
        \includesvg[width=\textwidth]{figs/main_fig/main_fig_pre_comm_and_gp.svg}
        \caption{Before inter-robot exchange}
        \label{subfig:main_fig:pre_comm}
    \end{subfigure}~
    \begin{subfigure}[t]{0.474\textwidth}
        \centering
        \includesvg[width=\textwidth]{figs/main_fig/main_fig_post_comm_w_connecting_factor.svg}
        \caption{After inter-robot exchange}
        \label{subfig:main_fig:pst_comm}
    \end{subfigure}~
    \caption{In \emph{DistGP}, each robot maintains a block of inducing points which parameterise their local GP (far-left) and into which observations can be fused. Communicating robots harmonise models by connecting inducing points via GP consistency factors and exchanging GBP messages (far-right). Here, the map at each point is predicted by the closest robot. Before communication (centre-left) the map is discontinuous and inaccurate, but this is resolved using only local inter-robot exchanges (centre-right).}
    \label{fig:main_fig}
    \vspace{-9pt}
\end{figure*}
Many mapping tasks involve large areas and dynamic maps, suggesting a divide-and-conquer approach using multiple robots. Centralised communication and computation is often not feasible in domains like environmental monitoring \cite{environmentalmonitoring}, robotic agriculture \cite{agriculture} or space exploration \cite{spaceexploration}, which involve many robots accumulating large datasets in poorly connected regions. Further, centralised approaches can be slow where robots may need up-to-date local models to guide exploration. These considerations motivate decentralised approaches, with local computation and mesh connectivity.

In this work, we present DistGP: a multi-robot algorithm to fit an arbitrary global function using only local observation and computation; with peer-to-peer communication. We use sparse GPs whose inducing points approximate a full GP at lower cost, and act as local summaries of the function. They provide a useful abstraction for data fusion and inter-robot exchange. Further, they are assumed to be Gaussian distributed, enabling decentralised inference with GBP. 

We first consider tree-structured GPs (TSGPs) \cite{bui2014tree}, whose inducing points are partitioned into sparsely connected blocks. Each block covers a distinct region of the input space and is maintained by a single robot. Robots exchange GBP messages with their neighbours to ensure their inducing points are consistent. TSGP restricts connectivity to a tree to guarantee rapid convergence, however we show empirically that this constraint limits accuracy: inducing points close in Euclidean space may not be directly connected, leading to discontinuities at their boundary. Further, in dynamic multi-robot scenarios, two communicating robots are prevented from sharing information if doing so would introduce a cycle.

To address this limitation, we propose a simple extension: relax the tree constraint and allow cycles. While GBP in our loopy model requires more iterations to converge, we demonstrate a substantial improvement in accuracy. Moreover, our approach is better suited to dynamic and asynchronous settings, where inter-robot communication patterns are continually changing -- unsupported by TSGP, whose message schedule requires fixed connectivity. Our loopy model naturally exploits all robot encounters and converges with asynchronous message passing. 
Our proposed loopy factor graph model is illustrated in Fig.~\ref{fig:main_fig}.

Our motivations are similar to Yu et al. \cite{yu2022dinno}, who use a distributed NN optimiser (DiNNO) for multi-robot learning.
In contrast to DiNNO, DistGP i) enforces inter-robot consistency locally, only where both robots have inducing points, rather than in global weight-space, ii) is a probabilistic model supporting on-the-fly data fusion for online learning, and iii) is scalable as each robot stores only a local map estimate.

In short our contributions are as follows:
\begin{enumerate}
    \item A new distributed GP model, trainable with GBP, which outperforms TSGPs \cite{bui2014tree}. 
    \item Empirical analysis showing the model can be built on-the-fly and admits distributed, asynchronous inference via GBP. It is thus well-suited to multi-robot mapping.
    \item Experimental evaluation of our method on environmental monitoring and occupancy mapping simulations. We show superior performance to DiNNO \cite{yu2022dinno}.
\end{enumerate}

\section{Related Work}
\paragraph{Distributed NN Maps} Many works build distributed implicit neural maps through inter-robot collaboration \cite{yu2022dinno,deng2024macim,asadi2024di,zhao2024distributed,zhao2025ramen}, either using consensus alternating direction method of multipliers (CADMM) \cite{admm} or direct regularisation to constrain NN weights to be similar. DiNNO \cite{yu2022dinno} is most similar to ours as it is a general method for multi-robot function learning rather than a specific method for distributed $3$D reconstruction. While distributed neural maps enforce agreement in global weight space, we represent maps with factorised GPs which enable local consistency constraints, specific to regions in which both robots have inducing points.

In DiNNO, each robot maintains a NN estimate of the global map, which are brought into agreement by CADMM. In this work however, we seek to build a distributed map: each robot maintains their local segment of the map and inter-robot interactions are used to ensure each segment is consistent with its neighbours. We argue DistGP is:
\begin{enumerate}
    \item {\bf more scalable.} For problems with large areas and large numbers of robots, each robot having a global map is inefficient and reaching consensus is more challenging.
    \item {\bf more modular.} We divide the map into segments which can be individually copied and communicated as required. In contrast, it is unclear how a map stored in the weights of a NN can be subdivided.
    \item {\bf also able to generate a shared global map.} In the extreme case that each robot shares its own inducing point posterior, and all others it has received thus far, we can achieve global map sharing as in DiNNO.
\end{enumerate}

\paragraph{Multi-Robot GPs} As in this work, \cite{luo2018adaptive} use a distributed GP for multi-robot learning. Each robot first trains a local GP on its observations and they are combined as a mixture of GPs, with location-varying weights. The Expectation Maximisation algorithm %
optimises the local models in tandem with the mixture weights of the different observations. Unlike our method, full connectivity is needed for normalised mixture weights, and \emph{all} robots must generate a prediction at any test point. In addition, the use of dense GPs precludes scaling to large datasets. \cite{habibi2021human} present a multi-agent inducing point GP method for human trajectory prediction, however they do not consider the problem of distributed learning of a function over the space being navigated.

\paragraph{Multi-Robot GBP} Recent work has shown GBP to be an effective distributed solver. Our work draws inspiration from \cite{robotweb}, in using GBP exchange between robots to enable collaborative estimation. However, \cite{robotweb} estimate the robot locations, where we estimate a distributed map. Like us, \cite{gbpstack} utilise GBP for multi-robot mapping. Each robot maintains a grid of the input space with one variable per cell to represent its function value. These are updated via i) observation by the robot or ii) inter-robot communication where two robots enforce consistency between their pairs of grid variables. Unlike DistGP, this method does not benefit from function approximation -- it is unable to generalise to unobserved regions and can only be queried at fixed resolution.

\section{Background}
\label{sec:background}
\subsection{Gaussian Process Regression}
\label{subsec:gpr}
A GP is a generalisation of a Gaussian distribution to the infinite dimensional space of functions. It is defined by a mean function $\meanfunction: \Reals^\inputdim\rightarrow \Reals$, and a covariance (or kernel) function $\kernelfn_\kernelhyps: \Reals^{\inputdim\times \inputdim}\rightarrow \Reals$, where $\inputdim$ is the input dimension. $\kernelfn_\kernelhyps(\x_a, \x_b)$ is the covariance between the function values at pairs of input locations $(\x_a,\x_b)$ and has hyperparameters $\kernelhyps$. 

For regression, a GP can be used as a prior $p(f(\cdot)|\theta)$ on the function to be fitted, $f: \Reals^D\rightarrow\Reals$. For an iid Gaussian likelihood on observations $\y=[y_1,\ldots,y_\numobs]^\transpose$ at inputs $\X=[\x_1^\transpose,\ldots,\x_\numobs^\transpose]^\transpose$, the posterior of $f(\cdot)$ at test point $\x^*$ is:\vspace{-2pt}
\begin{align}
    p\left(f(\x^*)|\y\right) &= \Normal\left(\mu_{*},\Sigma_{*}\right) \label{eq:gp_posterior}\\
    \mu_{*}&:=\kernelvector_{x^*x}\left(\kernelmatrix_{x} + \sigma^2 I_x\right)^{-1} \y,\nonumber\\
    \Sigma_{*}&:=\kernelmatrix_{x^*} - \kernelvector_{x^*x}\left(\kernelmatrix_{x} + \sigma^2 I_x\right)^{-1} \kernelvector_{xx^*}, \nonumber
\end{align}
where $\kernelmatrix_{x}\in\Reals^{\numobs\times\numobs}$ has $ij$th element $\kernelfn_\kernelhyps(\x_i,\x_j)$,
$\kernelmatrix_{x^*}\in \Reals^{1\times1}$ has element $\kernelfn_\kernelhyps(\x^*,\x^*)$, $\kernelvector_{xx^*}$ is a length $\numobs$ vector with $l$th element $\kernelfn_\kernelhyps(\x_l,\x^*)$,
and $\sigma$ is the observation noise. Hyperparameters $\theta$ and $\sigma$ can be trained by gradient ascent of the (log-)marginal likelihood, $p(\y|\theta)=\int p(\y|\f)p(\f|\theta)\calcd \f$.

The time complexity of the matrix inverse in $\eqref{eq:gp_posterior}$ is $O(N^3)$, which is prohibitive for large datasets, motivating approximations with cheaper training and prediction.

\subsection{Inducing Point Approximations}
Inducing point GPs seek to compress the dataset of $\numobs$ observations into $\numinducing<\numobs$ pseudo-datapoints (inducing points) which capture its important characteristics at reduced computational cost. One such method is the Fully Independent Training Conditional (FITC) approximation \cite{snelson2005sparse,csato2002sparse}. The inducing variables $\u\in\Reals^\numinducing$ approximate $f(\cdot)$ at the chosen input locations $Z=[\z_1^\transpose,\ldots,\z_\numinducing^\transpose]^\transpose$. The generative model is \vspace{-7pt}
\begin{align}
    q(\f,\u)=q(\f|\u)p(\u)= \prod_{i=1}^\numobs p(f_i|\u) p(\u),
\end{align}
where the conditional independence of $\f | \u$ is a model assumption which enables cheaper inference and learning, $p(\u)$ is a dense GP prior over the inducing points. The resulting posterior predictive $p\left(f(\x^*)|\y,X\right)$ requires inversion of dense $\numinducing\times \numinducing$ and diagonal $\numobs\times \numobs$ matrices, thus avoiding the $O(\numobs^3)$ cost of inverting a dense $\numobs\times \numobs$ matrix.

\subsection{Tree-Structured Gaussian Process Approximations}
While inducing point approximations such as FITC are efficient when $\numinducing<\!<\numobs$, Bui and Turner \cite{bui2014tree} argue that many datasets require $M\approx N$ for accurate prediction. For efficient inference in this setting, they propose approximating a dense prior over $\u$ by dividing them into blocks $\u=[\u_{B_1}^\transpose, \ldots, \u_{B_K}^\transpose]^\transpose$ and connecting them in a tree \vspace{-7pt}
\begin{align}
    q(\u) &= q(\u_{B_1})\prod_{k=2}^K p(\u_{B_k} | \u_{\parent(B_k)}), \label{eq:fact_prior}
\intertext{where $B_1$ is the root node and $\parent(B_k)$ is the parent of $B_k$. \endgraf Each block generates a disjoint observation subset $\f_{C_k}$\vspace{-7pt}}
    q(\f|\u)&=\prod_{k=1}^K q(\f_{C_k}|\u_{B_k}). \label{eq:fact_likelihood} 
\end{align}
with connections between elements of $\f$ and $\u$ determined by proximity. For example, $f_a:=f(\x_a)$ may be connected to the $\u_{B_k}$ whose input location centroid is closest to $\x_a$. By minimising the KL divergence from the dense joint $p(\f,\u)$ to approximation $q(\f,\u)$, they derive the form of factors $q(\u_{B_k} | \u_{\parent(B_k)})$ and $q(\f_{C_k}|\u_{B_k})$, showing they are equal to standard GP conditionals $p(\u_{B_k} | \u_{\parent(B_k)})$ and $p(\f_{C_k}|\u_{B_k})$.

As TSGP has a sparse set of factors $\{p(\u_{B_k}|\u_{\parent(B_k)})\}_k$ in place of a dense prior $p(\u)$, inference can be efficient in models where the number of inducing points is large. Bui and Turner \cite{bui2014tree} show how the posteriors $\{q(\u_{B_k}|\y)\}_k$ can be inferred with GBP message passing. The tree-structured topology ensures GBP converges in one sweep from leaves to root and another from root to leaves. %

After inference, prediction at a new $\x^*$ can be made by finding a nearby block of inducing points $\u_{B_{k^*}}$ (e.g. with input locations whose centroid is closest) and estimating $f(\x^*)$ by marginalising over its posterior $q\left(\u_{B_{k^*}}|\y\right)$:
\begin{align}
    q\left(f(\x^*)|\y\right) = \int p\left(f(\x^*)|\u_{B_{k^*}}\right)q\left(\u_{B_{k^*}}|\y\right) d\u_{B_{k^*}}.
    \label{eq:tsgp_predict}
\end{align}
This prediction relies only on the closest $\u_{B_{k^*}}$, avoiding test-time communication with other parts of the graph.

\subsection{Belief Propagation}
Belief propagation (BP) \cite{pearl2014probabilistic} is a message passing algorithm to compute the marginal beliefs of variables in a graphical model. Given the joint distribution's factorisation\vspace{-7pt}
\begin{align}
    p(\vargeneric_1, \ldots, \vargeneric_\numvars) &= \frac{1}{Z}\prod_{j} \factor_j (\{\vargeneric_l\}_{l\in\nei(j)}),
\intertext{BP computes (or estimates) $\{p(\vargeneric_j)\}_{j=1}^{\numvars}$. If factors $\factor_j$ are conditional on observations $\y$, then the marginals constitute a mean-field posterior $q(\vargeneric_1,\ldots,\vargeneric_n | \y) = \prod_{i=1}^n p(\vargeneric_i|\y)$. \endgraf Each edge in a factor graph stores both a factor-to-variable message $\messagefactovar{j}{i}(\vargeneric_i)$ and a variable-to-factor message $\messagevartofac{i}{j}(\vargeneric_i)$. BP iterates between updating these:}\vspace{-7pt}
    \messagefactovar{j}{i}(\vargeneric_i) &\leftarrow \!\!\!\!\!\!\!\!\!\sum_{\{\vargeneric_{r}\}_{\nei(j)\setminus i}} \!\!\!\!\!\!\!\!\factor_j(\{\vargeneric_l\}_{\nei(j)}) \!\!\!\!\!\prod_{k\in\nei(j)\setminus i}\!\!\!\!\!\!\messagevartofac{k}{j}(\vargeneric_k) \label{eq:bp_fac_to_var}\\
    \messagevartofac{i}{j}(\vargeneric_i) &\leftarrow \!\!\!\!\!\prod_{s\in\nei(i)\setminus j} \!\!\!\!\!\messagefactovar{s}{i}(\vargeneric_i), \label{eq:bp_var_to_fac}
\intertext{where $\nei(r)\setminus s$ is the set of all nodes connected to $r$ except $s$. %
After convergence, the marginals are given by the product of incoming message from all connected factors to a variable,}\vspace{-10pt}
    p(\vargeneric_i)&=\prod_{s\in\nei(i)} \messagefactovar{s}{i}(\vargeneric_i).
\end{align}
Message and belief updates are more efficient using the natural parameterisation of messages and factors, as the product of messages reduces to the sum of their parameters.

BP is only guaranteed to converge to exact marginals in tree-structured models, where convergence can be reached by a leaves-to-root then root-to-leaves sweep of message updates. Despite a lack of guarantees, BP often performs well on tasks involving cyclic graphs \cite{gallager1962low,murphy2013loopy}. 

\subsubsection{Gaussian Belief Propagation}
For Gaussian models, the natural parameters are the precision (inverse covariance) $\Lambda:=\Sigma^{-1}$ and information vector $\eta:=\Sigma^{-1}\mu$. Factor-to-variable updates \eqref{eq:bp_fac_to_var} take the form (see \cite{davison2019futuremapping})
\begin{align*}
    \infomessage{\factor_j}{\vargeneric_i} &\leftarrow \!\info^{(\factor_j)}_{i} \!\!-\!  \Prec^{(\factor_j)}_{i,\setminus i} \left(\Prec^{(\factor_j)}_{\setminus i, \setminus i} + \Precmessage{\x_{\setminus i}}{\factor_j} \!\right)^{-1} \!\!\!\left(\eta^{(\factor_j)}_{\setminus i} \!\!+ \infomessage{\x_{\setminus i}}{\factor_j}\right)\\
    \Precmessage{\factor_j}{\vargeneric_i} &\leftarrow \Prec^{(\factor_j)}_{i,i} -  \Prec^{(\factor_j)}_{i,\setminus i} \left(\Prec^{(\factor_j)}_{\setminus i, \setminus i} + \Precmessage{\x_{\setminus i}}{\factor_j} \right)^{-1} \Prec^{(\factor_j)}_{\setminus i,i}
\end{align*}
where $(\info^{(\factor_j)}, \Prec^{(\factor_j)})$ are parameters of $\factor_j$ and subscript $_{\vargeneric_{\setminus i}\rightarrow\factor_j}$ denotes concatenation of all messages incoming to $\factor_j$ except that from $\vargeneric_i$. Variable-to-factor messages \eqref{eq:bp_var_to_fac} are 
\begin{align*}
\infomessage{\vargeneric_i}{\factor_j}\leftarrow\!\!\!\!\!\sum_{k\in\nei(i)\setminus j}\!\!\!\infomessage{\factor_k}{\vargeneric_i}, \quad \Precmessage{\vargeneric_i}{\factor_j}\leftarrow\!\!\!\!\!\sum_{k\in\nei(i)\setminus j}\!\!\!\Precmessage{\factor_k}{\vargeneric_i}.
\end{align*}

As in the general case, convergence in loopy Gaussian models is not guaranteed, though if GBP \emph{does} converge in such cases, marginal means are guaranteed to be correct \cite{weiss1999correctness}.

\section{Method}
\subsection{Problem Statement}
We aim to learn a scalar map $f: \Reals^D\rightarrow\Reals$ distributed between $K$ robots, $f_\mathrm{dist}=\{f_{k}\}_{k=1}^K$, $f_\mathrm{dist}(\x^*)=f_i(\x^*) \iff\x^*\in\mathcal{R}_i$, where partial map $f_i(\cdot)$ is maintained by robot $i$ which covers region $\mathcal{R}_i$. For static maps, $D=D_\mathrm{space}$, where dynamic maps are also a function of time: $D=D_\mathrm{space}+1$.

\subsection{Improving Over Tree-Structured GPs}
\label{subsec:improving_tsgp}
\begin{figure}[t!]
    \centering
     \begin{subfigure}[t]{0.22\textwidth}
        \centering
        \includegraphics[width=\textwidth]{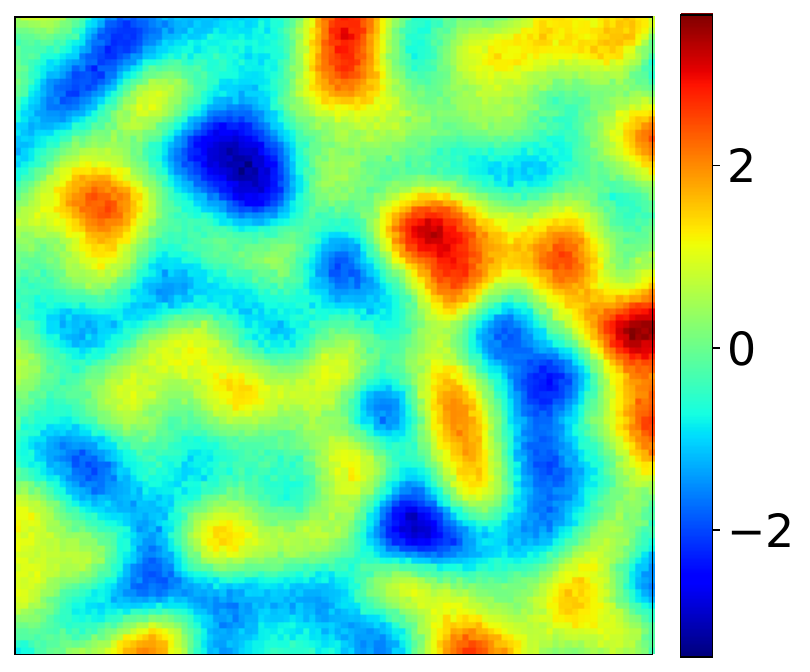}
        \caption{Ground truth $+$ noise}
        \label{subfig:tspgp_analysis:ground_truth}
    \end{subfigure}~
    \begin{subfigure}[t]{0.22\textwidth}
        \centering
        \raisebox{-3pt}{\includegraphics[width=\textwidth]{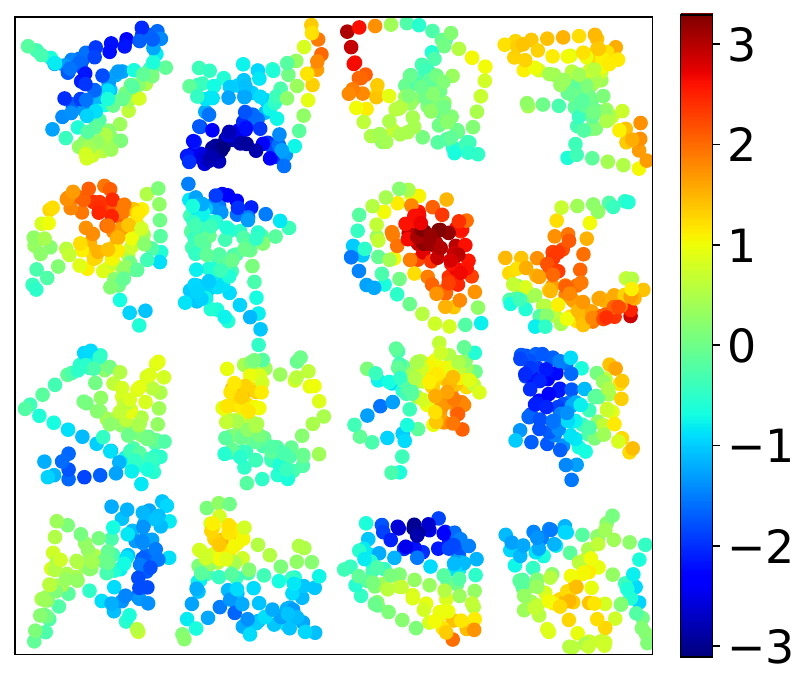}}
        \vspace*{-12.2pt}
        \caption{Training data}
        \label{subfig:tspgp_analysis:training_data}
    \end{subfigure}\\
    \begin{subfigure}[t]{0.22\textwidth}
        \centering
        \includegraphics[width=\textwidth]{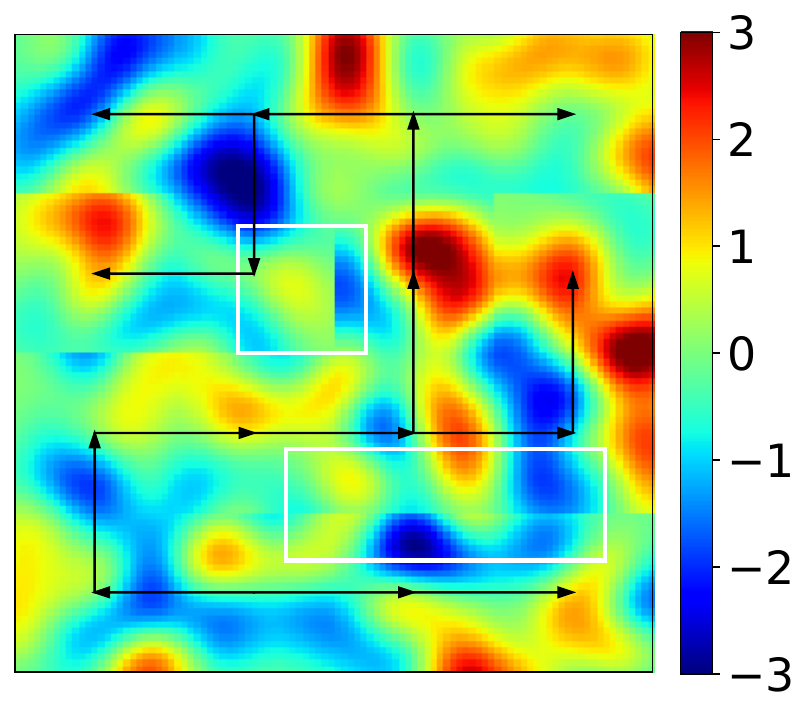}
        \caption{TSGP}
        \label{subfig:tspgp_analysis:tsgp_preds}
    \end{subfigure}~
    \begin{subfigure}[t]{0.22\textwidth}
        \centering
        \includegraphics[width=\textwidth]{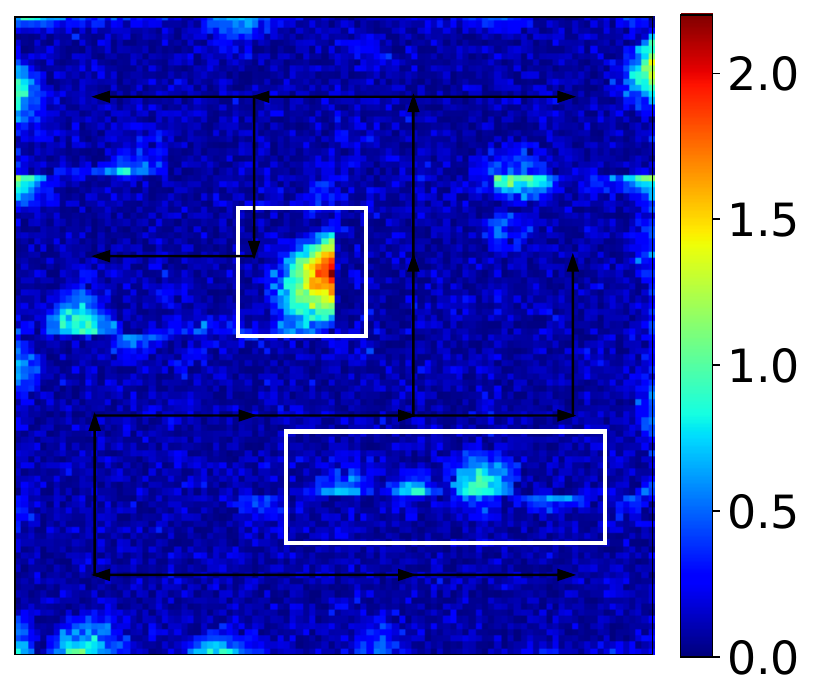}
        \caption{TSGP Abs. Error}
        \label{subfig:tspgp_analysis:tsgp_error}
    \end{subfigure}\\
    \begin{subfigure}[t]{0.22\textwidth}
        \centering
        \includegraphics[width=\textwidth]{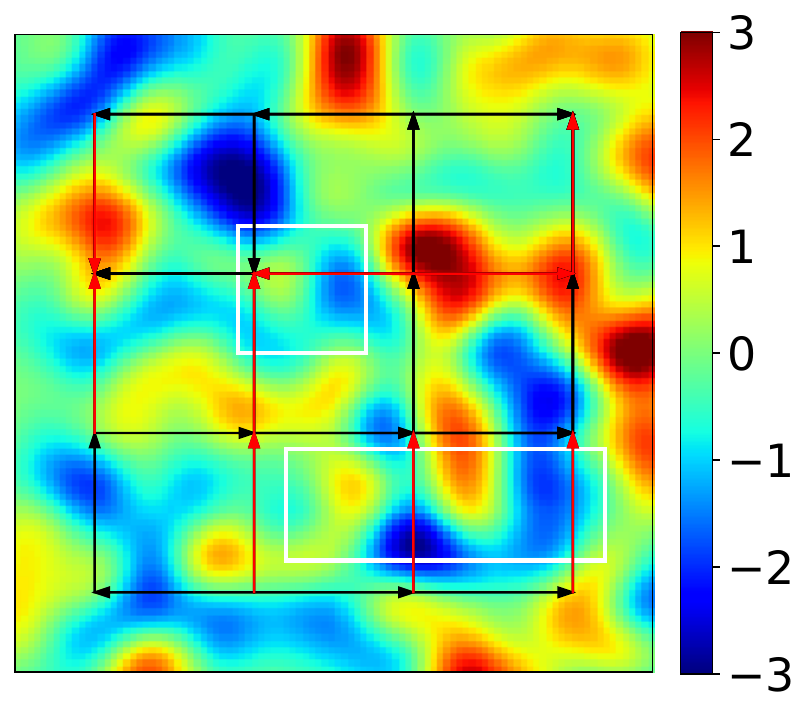}
        \caption{TSGP + $9$ Edges}
        \label{subfig:tspgp_analysis:tsgp_extra_preds}
    \end{subfigure} ~
    \begin{subfigure}[t]{0.22\textwidth}
        \centering
        \includegraphics[width=\textwidth]{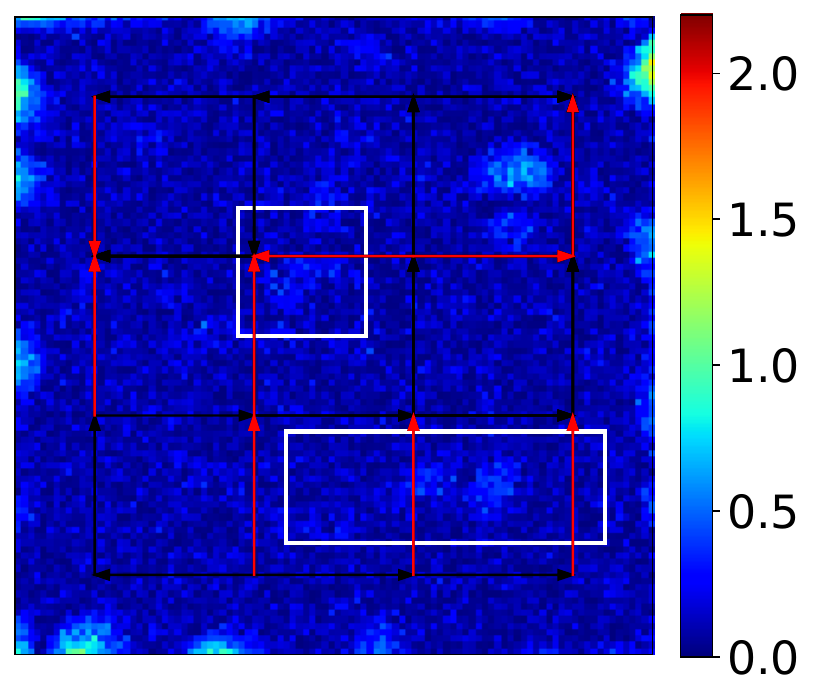}
        \caption{TSGP + $9$ Edges Abs. Error}
        \label{subfig:tspgp_analysis:tsgp_extra_error}
    \end{subfigure}
    \caption{{\bf Limitations of TSGP.} Some robots remain unconnected to ensure tree connectivity (as in \subref{subfig:tspgp_analysis:tsgp_preds}). This leads to discontinuities in the prediction at the boundary between unconnected robots (\subref{subfig:tspgp_analysis:tsgp_error}), which can be reduced significantly by adding loop-forming edges (\subref{subfig:tspgp_analysis:tsgp_extra_error}) (see white rectangles).}
    \label{fig:tsgp_analysis}
    \vspace{-9pt}
\end{figure}

\begin{figure}[t]
    \centering
    \includegraphics[width=0.239\textwidth]{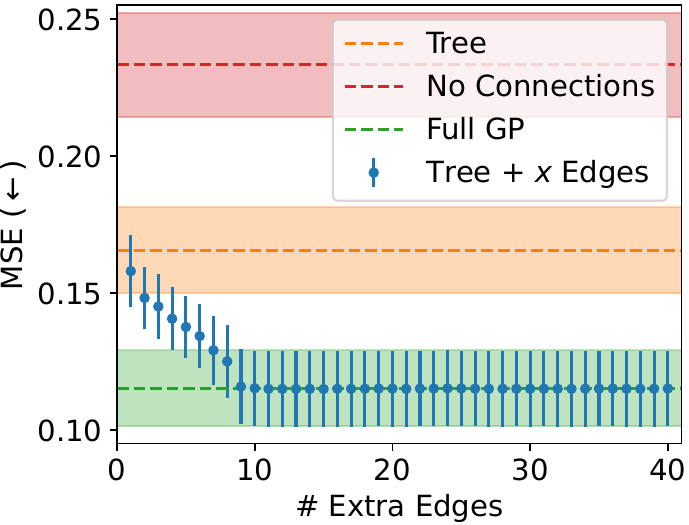}
    \includegraphics[width=0.239\textwidth]{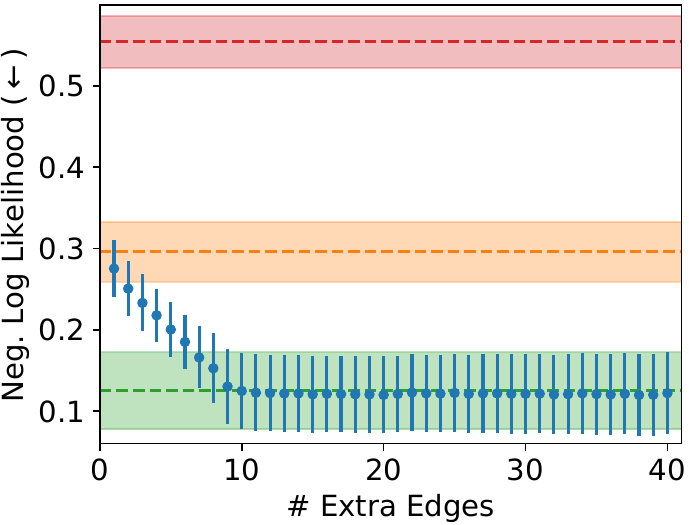}
    \caption{{\bf Benefit of additional edges.} The prediction error (vertical axis) reduces as more edges are added to the TSGP model (horizontal axis, $x=0$ is TSGP), but plateaus after $\sim9$ extra edges, where accuracy is equal to a dense GP prior on all inducing points (green). Results averaged over $10$ seeds, errorbars cover $\pm1$ standard error (SE) around mean.}
    \label{fig:acc_vs_loops}
    \vspace{-10pt}
\end{figure}
The following features TSGP make it a good candidate for distributed multi-robot mapping:
\begin{enumerate}
	\item Each inducing block $\u_{B_k}$ models a different subset of observations \eqref{eq:fact_likelihood}, so robots can update their local posteriors $q(\u_{B_k}|\y_{C_k})$ with new experience without needing to communicate with others.
	\item Each $\u_{B_k}$ only connects to its neighbours \eqref{eq:fact_prior}. With GBP inference, this ensures inter-robot messages are local exchanges -- no global communication is needed.
\end{enumerate}

To evaluate TSGP, we train it on simulated multi-robot data. We sample a static, ground truth function from a GP with $D=2$, which the robots observe (with additive Gaussian noise) at each timestep. The input space covering $[-1, 1]\!\times\![-1, 1]$ is divided into a $4\!\times\!4$ grid, with a robot occupying each cell. We generate robot trajectories by randomly sampling a target
within each cell, which the robot heads towards at a constant speed. Once within $\epsilon=0.02$, the target is resampled. The resulting observations are presented in Fig.~\ref{subfig:tspgp_analysis:training_data}. Each robot's inducing points are placed on a regular grid spanning its cell. Blocks of inducing points between robots in adjacent cells are connected, such that the graph of all inducing point blocks forms a tree (black arrows in Fig.~\ref{subfig:tspgp_analysis:tsgp_preds}). The model is batch-trained on the full dataset.

The resulting predictions and error (Figs.~\ref{subfig:tspgp_analysis:tsgp_preds} and~\ref{subfig:tspgp_analysis:tsgp_error}) show discontinuities in the learnt function and large errors at the boundaries between unconnected cells. This pathology results from the model's tree-structure: a tree connecting nodes on a grid with $D\geqslant2$ will necessarily include unconnected neighbours. The learnt function is discontinuous at their boundary due to an absence of inter-robot factors.

To quantify this effect, we train a sequence of models, starting at TSGP and adding edges one-by-one, prioritising those between robots close in Euclidean space but distant in the graph. We also train two baselines: i) with no inter-robot connections (each robot fits an independent FITC GP), and ii) with a full GP prior over \emph{all} inducing points. We train each model $10$ times on different datasets, each randomly sampled as in Fig.~\ref{subfig:tspgp_analysis:training_data}. We evaluate on a dense grid, but remove test points less than $d=0.06$ from any training observations.

The results (Fig.~\ref{fig:acc_vs_loops}) show that models with more edges have higher accuracy, until $\sim\!9$ additional edges after which performance matches the full GP. Examining predictions and errors for a model with $9$ extra edges on the example dataset (Figs.~\ref{subfig:tspgp_analysis:tsgp_extra_preds} and~\subref{subfig:tspgp_analysis:tsgp_extra_error}) we see discontinuities are much reduced (e.g. in white rectangles, Figs.~\ref{subfig:tspgp_analysis:tsgp_error} and~\ref{subfig:tspgp_analysis:tsgp_extra_error}). We believe performance plateaus at $9$ extra edges (red arrows, Fig.~\ref{subfig:tspgp_analysis:tsgp_extra_preds}) because here each node connects to its immediate NESW neighbours. Further longer-range edges have little impact. This is consistent with repeated analyses on $3\!\times\!3$ and $5\!\times\!5$ grids, which plateau at $4$ and $16$ extra edges respectively.

\subsection{Asynchronous Model Construction and Training}
\label{subsec:async_model_construction}
We have shown that TSGP's mapping performance can be improved with additional connections. However, we have thus far assumed that all data are available upfront, and any robot could be connected to any other. In multi-robot mapping these assumptions are not realistic, so we now design a model which can be trained incrementally, with dynamic connectivity as per the robots' relative locations. We will now describe (1) how observations are fused into local models, %
(2) where new inducing points are added, %
(3) how inter-robot connections are made, (4) how kernel hyperparameters are learnt and (5) how predictions are generated. %

\subsubsection{Adding New Observations}
\label{subsec:new_observations}
Desirable features of a robot's local model include i) scalability w.r.t the number of observations, ii) ability to efficiently incorporate new observations, iii) ability to discard old observations after model update. To this end, we employ the FITC GP approximation \cite{snelson2005sparse,csato2002sparse} (see factor graph in Fig.~\ref{subfig:main_fig:pre_comm}). For a robot with inducing points $\u_{B_i}$ and observations $\y^{(C_i)}=[y_{1}^{(C_i)}, \ldots, y_{N_i}^{(C_i)}]^\transpose$ at $\X^{(C_i)}=[(\x_{1}^{(C_i)})^\transpose,\ldots,(\x_{N_i}^{(C_i)})^\transpose]^\transpose$, this approximation can be written\vspace{-6pt}
\begin{align}
    q(\y_i, \f_{C_i}|\u_{B_i}) &= \prod_{n=1}^{N_i} p(y_{n}^{(C_i)} | f_{n}^{(C_i)}) p(f_{n}^{(C_i)} | \u_{B_i}).
\end{align}

For $\numobs_i < \numinducing_i$, this approximation has inference complexity $O\left(\numinducing_i^2\numobs_i\right)$, and is therefore scalable to large $\numobs_i$. Further, a new observation $y_{\numobs_i+1}^{(C_i)}$ can be easily added online by creating a $f_{N_i + 1}^{(C_i)}$ variable, new $p(y_{N_i + 1}^{(C_i)} | f_{N_i + 1}^{(C_i)})$, $p(f_{N_i + 1}^{(C_i)} | \u_{B_i})$ factors, and calculating the resulting message to variable $\u_{B_i}$. Last, individual observations can be discarded after their message from factor $p(f_{n}^{(C_i)} | \u_{B_i})$ to $\u_{B_i}$ has been computed. We maintain a unary factor on $\u_{B_i}$ to accumulate messages from discarded observations. After this, factors $p(y_{n}^{(C_i)} | f_{n}^{(C_i)})$, $p(f_{n}^{(C_i)} | \u_{B_i})$ and variable $f_{i,m}$ are removed.

Discarding observation factors online introduces two sources of approximation relative to batch fitting. First, in most cases we add inducing points incrementally, as the robot gathers experience. Thus inducing points which have been added after an observation factor has been discarded will not benefit from the corresponding message. This does not affect inference when the inducing point locations are fixed throughout. Second, for non-linear observation factors $p(y_{n}^{(C_i)} | f_{n}^{(C_i)})$, the message to $\u_{B_i}$ depends on the estimate of $f_{n}^{(C_i)}$. Observations added after $y_{n}^{(C_i)}$ is removed could have affected $\mathbb{E}[f_{n}^{(C_i)}]$ and therefore the message to $\u_{B_i}$.

\subsubsection{Adding New Inducing Points}
\label{subsec:new_inducing}
We efficiently add inducing points to the model using the \emph{greedy variance selection} method of \cite{burt2020convergence}. This method i) chooses an inducing point at the observation location with the highest marginal output variance under $p(f_n^{(C_i)}|\u_{B_i})$, ii) adds it to $Z_{B_i}$, iii) updates the model and iv) repeats until $\numinducing_i$ is reached.

In our setting, the set of observations grows with time; we do not have the static dataset assumed in \cite{burt2020convergence}. Nonetheless we employ greedy variance selection in an online fashion, adding a new inducing point to the model every $k$ timesteps. We restrict our choice of new locations to observations which have not yet been fused and discarded (see Sec.~\ref{subsec:new_observations}). This ensures that the new inducing point will benefit from a strong message from the co-located observation.

In addition to selecting inducing points from observations, we find adding inducing points near the boundary between connecting robots provides the model more flexibility to make their representations consistent. To this end, we use two strategies: i) each robot adds a line of inducing points along the boundary; or ii) each robot adds inducing points at the same location as a subset of the other's near the boundary.

\subsubsection{Adding New Connections}
\label{subsec:new_connections}
We next consider how two communicating robots should exchange information to synchronise their representations. We connect their inducing points via a shared factor, but what form should this factor take? 
We first observe that the optimal prior over inducing points, i.e. that which most closely approximates a full GP over observations, is itself a full GP over the inducing points:
\begin{align}
    p(\u_{B_1},\ldots,\u_{B_b}) &= \Normal(\mathbf{0},\kernelmatrix_{\Zall}) \label{eq:full_gp_prior}\\
\intertext{where $K_{\Zall}\in \Reals^{\numinducing\times\numinducing}$ is the kernel matrix over all inducing points. This joint prior can be factorised sequentially:}
    p(\u_{B_1},\ldots,\u_{B_b}) &= p(\u_{B_1})\prod_{i=2}^b p(\u_{B_i} | \{\u_{B_j}\}_{j<i}) \label{eq:full_gp_prior_factorised}
\end{align}
meaning it can be distributed s.t. each robot ``owns'' one factor. However, this fully connected structure is computationally infeasible: inference scales as $O(\numinducing^3)$, where $\numinducing:=\sum_i \dim(\u_{B_i})$. Moreover, in realistic multi-robot settings, each robot can only communicate with nearby peers, where this factorisation implies all-to-all communication.

Despite these constraints, \eqref{eq:full_gp_prior_factorised} provides a theoretical target: it implies an inter-robot communication protocol which would converge to the ideal $p(\u_{B_1},\ldots,\u_{B_b})$ in the limit of full connectivity. The protocol is as follows:
\begin{enumerate}
    \item {\bf Initialisation:} To begin, each robot has a unary GP prior on its inducing variables $p(\u_{B_i})=\Normal(\mathbf{0},\kernelmatrix_{Z_{B_i}})$, equal to the message that $\u_{B_i}$ would receive from a global dense prior over inducing points \eqref{eq:full_gp_prior}, due to the marginal consistency of GPs. This ensures message updates are well-conditioned, enabling inference in each robot's local model before any inter-robot contact.
    \item \label{point:inter_robot_exchange}{\bf Inter-robot Exchange:} Suppose robot $i$ now encounters $j$. The two compare their indices to determine which will act as the parent in the connecting factor. If $i>j$, for example, robot $j$ will be the parent whose inducing points will condition $i$'s prior. Three updates are now executed: i) the unary prior is amended to a conditional: $p(\u_{B_i})\rightarrow p(\u_{B_i}|\u_{B_j})$ and connected to $\u_{B_j}$, ii) variable-to-factor messages $\message_{\u_{B_i}\shortrightarrow p(\u_{B_i}|\u_{B_j})},\,\, \message_{\u_{B_j}\shortrightarrow p(\u_{B_i}|\u_{B_j})}$  are updated, and iii) factor-to-variable messages $\message_{p(\u_{B_i}|\u_{B_j})\shortrightarrow\u_{B_i}},\,\, \message_{p(\u_{B_i}|\u_{B_j})\shortrightarrow\u_{B_j}}$ are updated.
    \item {\bf Decoupling:} The message exchanges in~\ref{point:inter_robot_exchange} are executed every timestep the two robots are within range. After separation, the child $i$ stores the previous state of $\message_{\u_{B_j}\shortrightarrow p(\u_{B_i}|\u_{B_j})}$ and $p(\u_{B_i}|\u_{B_j})$. The factor may be further conditioned on another robot's inducing variables later. A unary factor with value $\message_{p(\u_{B_i}|\u_{B_j})\shortrightarrow\u_{B_j}}$ is added to $\u_{B_j}$. 
\end{enumerate}

In the limit of all-to-all communication, we highlight that this routine constitutes inference in a model with a full GP prior \eqref{eq:full_gp_prior_factorised}. However, there are two sources of approximation in our approach. First, robots which are always far away and never make contact may never be connected via a shared edge. These robots are treated as independent. Second, robots which connect and then move out of range will be propagating stale messages. Thus, the asynchronous nature of the connections may limit performance. We empirically evaluate the impact of these approximations in Sec.~\ref{subsec:async_model_evaluation}.

\subsubsection{Distributed Hyperparameter Learning}
\label{subsec:distributed_gp_hyperparam_tuning}
Kernel hyperparameters $\theta$ are typically tuned by gradient ascent of the log-marginal likelihood (see Sec.~\ref{subsec:gpr}). Bui and Turner  \cite{bui2014tree} propose a message-passing routine to update kernel hyperparameters in TSGP. However, their method requires a full up-down sweeps of the tree to compute the gradient before it is applied. This is not applicable in our case, with cyclic models and sporadic connectivity. Instead, we compute stochastic, approximate marginal likelihood gradients based on each robot's individual experience, and use this to update their local copy of the kernel hyperparameters. We interleave hyperparameter tuning and GBP, updating the GP factors using the latest hyperparameters at each stage.  

\subsubsection{Prediction}
Prediction at a test point $\x^*$ can be made directly from the posterior of a chosen block of inducing points as in \eqref{eq:tsgp_predict}. The choice of which block $q(\u_{B_i}|\y)$ for a query $\x^*$ should ideally be consistent with how observations are connected to inducing points during training. However, this will depend on the application and where $q(\u_{B_i}|\y)$ are stored at prediction time. For DistGP, we explore two methods for mapping test queries to robots: i) predict at $\x^*$ with the robot whose current location $\x_{r_i,t}$ is closest, $r^*=\argmin_{r_i\in\{r_1,\ldots,r_R\}}||\x^*-\x_{r_i,t}||^2$ and ii) pre-allocate the input space regions $\{\mathcal{R}_i\}$ in which each robot is responsible for mapping and predicting: $r^*=r_i\iff \x^*\in \mathcal{R}_i$.

\subsection{Evaluation of Asynchronous Model Construction}
\label{subsec:async_model_evaluation}
To quantify the impact of asynchronous model construction and GBP (as described in Sec.~\ref{subsec:async_model_construction}), we compare its performance to a batch-fitted model with a full GP prior over all inducing points. We generate the trajectories and observations as in Sec.~\ref{subsec:improving_tsgp}, and again place inducing points on a regular grid within each robot's cell. However, in this case the robots learn online, incorporating new observations, connecting and exchanging messages as per Sec.~\ref{subsec:async_model_construction}. Robots can only communicate with others in a $\dcomm=0.15$ radius of their current location. This distance is a small fraction of the overall input space covering $[-1,1]\!\times\![-1,1]$. For comparison, we fit the batch models every $400$ steps, using all observations made so far. We evaluate predictions in six randomly chosen $0.3\!\times\!0.3$ subregions which are ``held out'': robots may move through these regions but do not make observations when inside them.

\begin{figure}[t]
    \centering
    \includegraphics[height=0.22\textwidth]{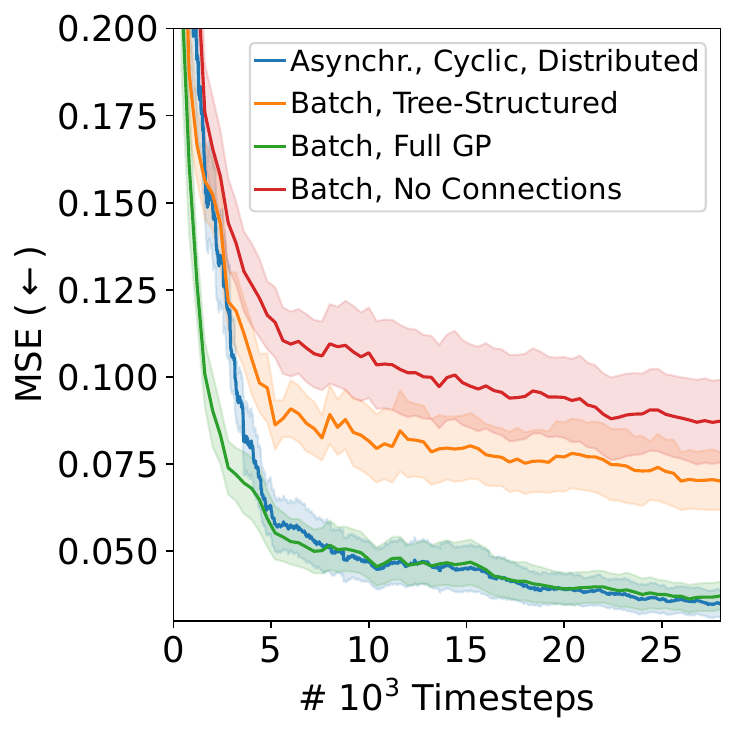}
    \includegraphics[height=0.216\textwidth]{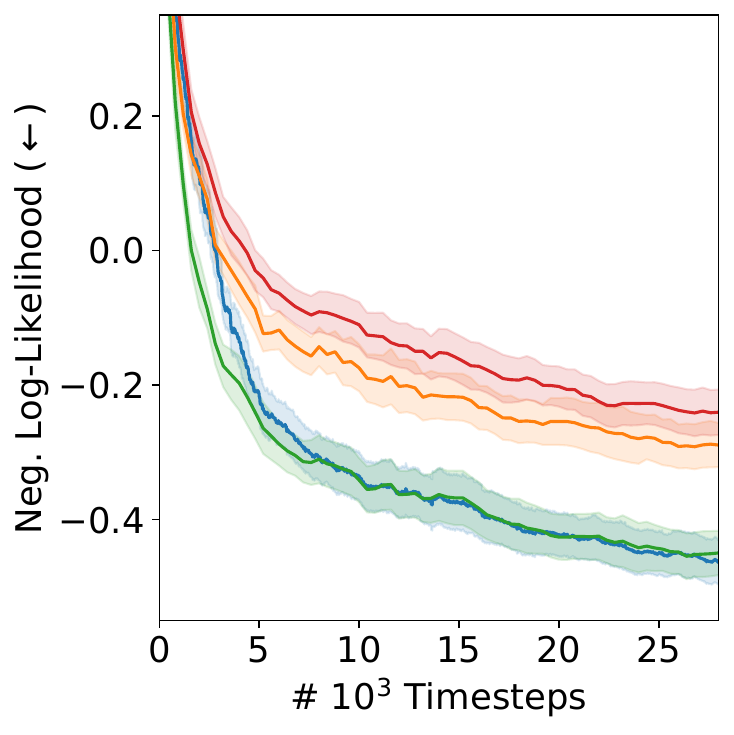}

    \caption{{\bf Asynchronous + distributed vs batch + centralised model.} We compare i) online learning, ad-hoc inter-robot connections, ad-hoc GBP (blue) against ii) a centralised implementation with synchronous GBP sweeps on batch data (red, orange, green). The former converges to the same performance as the centralised, dense prior (green). Averages over $10$ seeds, error bars cover $\pm1$ SE around the mean.}
    \label{fig:batch_vs_async}
    \vspace{-10pt}
\end{figure}

The results in Fig.~\ref{fig:batch_vs_async} show that while the distributed implementation (blue) is slower to converge, it achieves the same result as the batch-fitted dense GP prior over all inducing points (green). Further, the distributed implementation is able to rapidly achieve performance superior to a batch-fitted TSGP (orange) and a batch fitted model without connections (red). 
These results suggest we can maintain accuracy while learning a distributed map asynchronously.

\section{Experiments}
\label{sec:experiments}
We now evaluate DistGP, testing it on two distributed mapping simulations: sea-surface temperature (SST) prediction and occupancy mapping. We compare against DiNNO \cite{yu2022dinno}.

\subsection{Dynamic Sea Surface Temperature Tracking}
Many environmental monitoring applications require dynamic maps over large areas, in regions with poor connectivity. These challenges motivate a decentralised multi-robot mapping approach like ours. To evaluate our method in this setting, we simulate a multi-robot SST monitoring problem. The Optimally Interpolated SST (OISST) dataset~\cite{oisst,oisst2} combines satellite, buoy and ship measurements, and interpolates them on a dense grid of $1/4\textdegree$ lat/lon resolution at daily frequency. We select $5$ disjoint ocean regions, each covering a $4400\!\times\!4400$km$^2$ area, and divided into a $5\times5$ grid. We simulate $25$ aquatic robots spread over the region, one per cell, which make pointwise SST measurements as they move around. We assume the robots move at a low speed of $7$km/h, can communicate with others within $\dcomm\!\!=\!\!1500$km and observe the temperature every $3$ hours. As we only have access to daily mean temperatures, we assume the field remains constant throughout each day.

The aim is for the combined robot predictions over the area to accurately match the ground truth map at any given point in time. As the robots move slowly and the map evolves with time, their observation coverage is limited. To do well, therefore, requires robots are able to exchange useful information with their peers to improve their predictions in less recently observed regions. Combined predictions are generated on a dense grid, with each robot predicting the SST at all test points inside their cell.

We train our method by adding observations online and incorporating their messages into the inducing point posterior as per Sec.~\ref{subsec:async_model_construction}. Each robot retains $5$ observations in their model after which we fuse them into the unary factor on their inducing variables. Every other step, one of these observations is selected as a new inducing point via greedy variance selection. While this amounts to a relatively dense set of inducing points, we only retain the $200$ most recent inducing points, as older inducing points become irrelevant to current predictions as the field evolves. When two robots in adjacent cells connect, they both add a line of inducing points close to their shared border to provide extra flexibility to make their fitted functions consistent. We use the product of two Mat\'{e}rn-$1/2$ kernels as our covariance function, one over $2$D space and one over time; and we tune kernel $\kernelhyps$ as well as the observation noise $\sigma$ with the distributed method described in Sec.~\ref{subsec:distributed_gp_hyperparam_tuning}.

\begin{table}
\centering
    \begin{tabular}{ccc}
    \toprule
        {\bf Region} & {\bf DistGP (Ours)} & {\bf DiNNO} \\\midrule
        Central Atlantic & $\bm{0.134\pm0.001}$ & $0.175\pm0.001$\\
        North Indian Ocean & $\bm{0.160\pm0.001}$ &  $0.222\pm0.001$\\
        South Indian Ocean & $\bm{0.230\pm0.001}$ & $0.316\pm0.002$\\
        West Pacific & $\bm{0.108\pm0.001}$ & $0.166\pm0.001$\\
        East Pacific & $\bm{0.240\pm0.002}$ & $0.303\pm0.001$\\\bottomrule
    \end{tabular}
    \caption{{\bf SST prediction MSE.} Error in predicted temperatures over the map, averaged over all timesteps after a $25$ day ``burn-in''. Intervals cover $\pm$ SE over $4$ seeds.}
    \label{tab:sst_results}
    \vspace{-5pt}
\end{table}
\begin{table}
\centering
    \begin{tabular}{ccc}
    \toprule
        {\bf Comm. Setting} & {\bf DistGP (Ours)} & {\bf DiNNO} \\\midrule
        Default & $\bm{0.134\pm0.001}$ & $0.175\pm0.001$\\
        \arrayrulecolor{gray}\midrule
        Comm. Interval $1\rightarrow5$ & $\bm{0.138\pm0.001}$ & $0.249\pm0.001$\\
        Comm. Interval $1\rightarrow10$ & $\bm{0.143\pm0.001}$ & $0.294\pm0.001$\\
        \midrule \arrayrulecolor{black}
        $\dcomm$ $1500\mathrm{km}\rightarrow1100\mathrm{km}$ & $\bm{0.143\pm0.001}$ &  $0.420\pm0.005$\\
        $\dcomm$ $1500\mathrm{km}\rightarrow700\mathrm{km}$ & $\bm{0.169\pm0.001}$ &  $-^\dagger$\\
        \bottomrule
    \end{tabular}
    \caption{{\bf Varying communication.} MSE with different communication settings for the Central Atlantic region. $^\dagger$optimisation diverged. Intervals cover $\pm$ SE over $4$ seeds.}
    \label{tab:sst_comm}
    \vspace{-10pt}
\end{table}

The base model for DiNNO is a $6$-layer MLP with width $64$ and leaky-ReLU activations. At each timestep we train these networks using minibatch samples from the previous two weeks observations. This window length was found to perform best during validation.
Hyperparameters for both methods were chosen based on simulations in the Central Atlantic region from $2022/01/01$ to $2022/07/01$. We evaluate global map accuracy on a dense grid at every $3$ hour timestep between $2022/07/01$ and $2023/03/15$, following a $25$ day burn-in period.

The results (Tab.~\ref{tab:sst_results}), show that DistGP significantly outperforms DiNNO for all regions, suggesting more effective information sharing. We further examine the robustness of both methods under different communication settings: i) less frequent (every $5$ or $10$ steps instead of every step), and ii) shorter range ($\dcomm\in\{700\mathrm{km},1100\mathrm{km}\}$ instead of $1500$km). In both cases, we find DiNNO suffers considerable degradation in accuracy where DistGP performs similarly to the more frequent, longer-range communication.

\subsection{Online Occupancy Mapping}

\begin{figure*}[t!]
    \centering
     \begin{subfigure}[t]{0.24\textwidth}
        \centering
        \includegraphics[width=\textwidth]{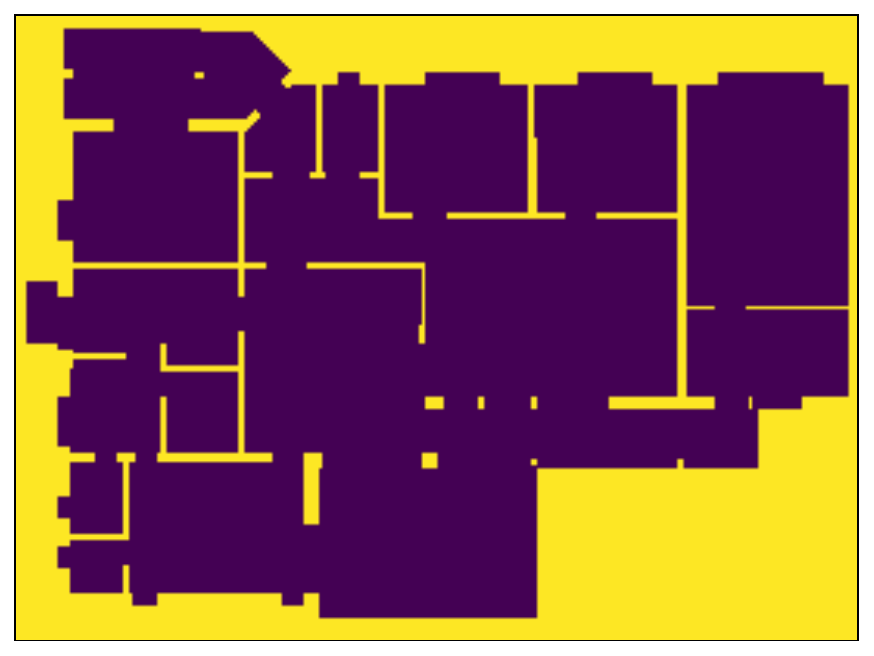}
        \caption{Ground truth}
        \label{subfig:occupancy_mapping:ground_truth}
    \end{subfigure}%
    \begin{subfigure}[t]{0.24\textwidth}
        \centering
        \includegraphics[width=\textwidth]{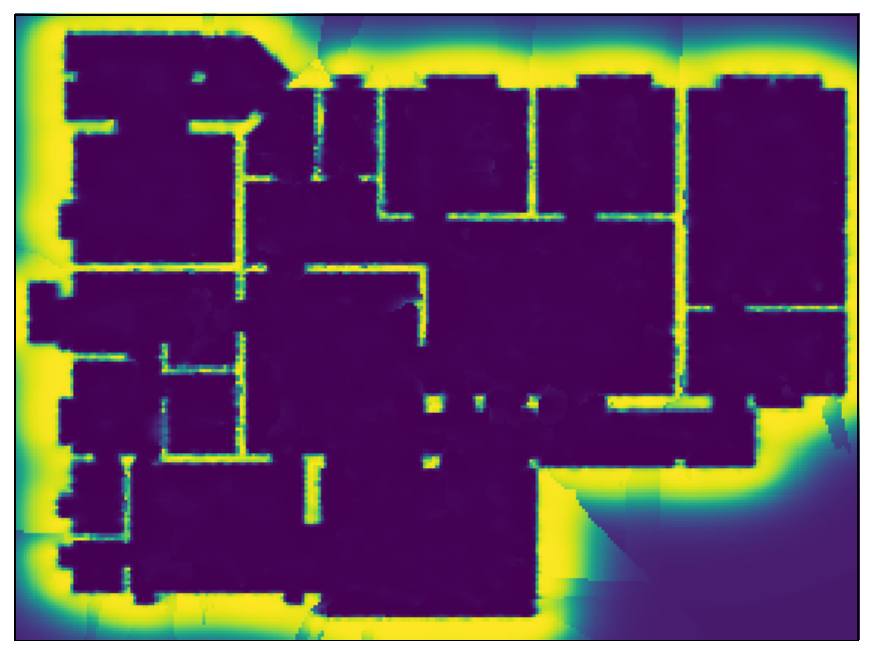}
        \caption{DistGP (ours), single pass}
        \label{subfig:occupancy_mapping:gp}
    \end{subfigure}
    \begin{subfigure}[t]{0.24\textwidth}
        \centering
        \includegraphics[width=\textwidth]{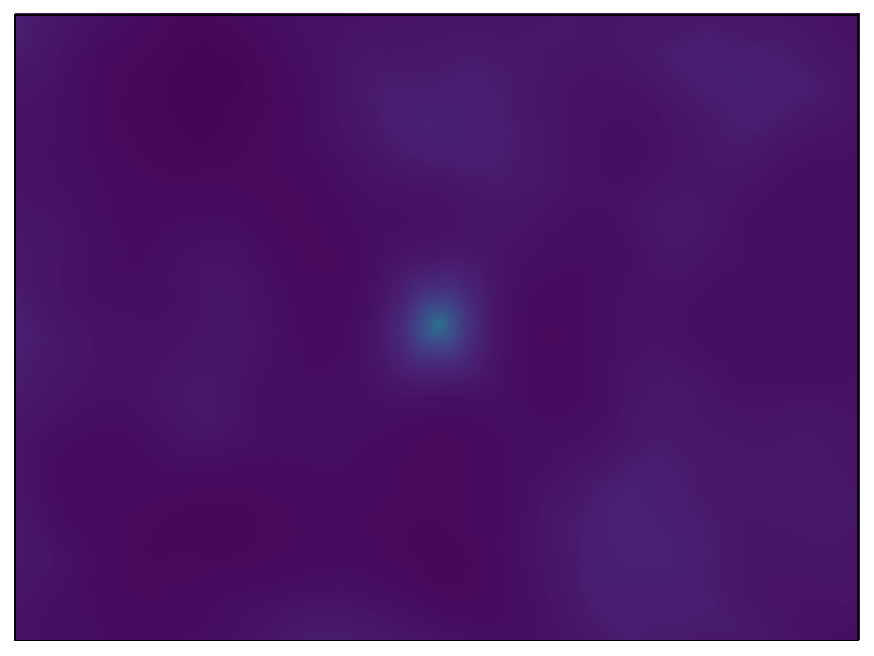}
        \caption{DiNNO, single pass}
        \label{subfig:occupancy_mapping:dinno_one_epoch}
    \end{subfigure} 
    \begin{subfigure}[t]{0.24\textwidth}
        \centering
        \includegraphics[width=\textwidth]{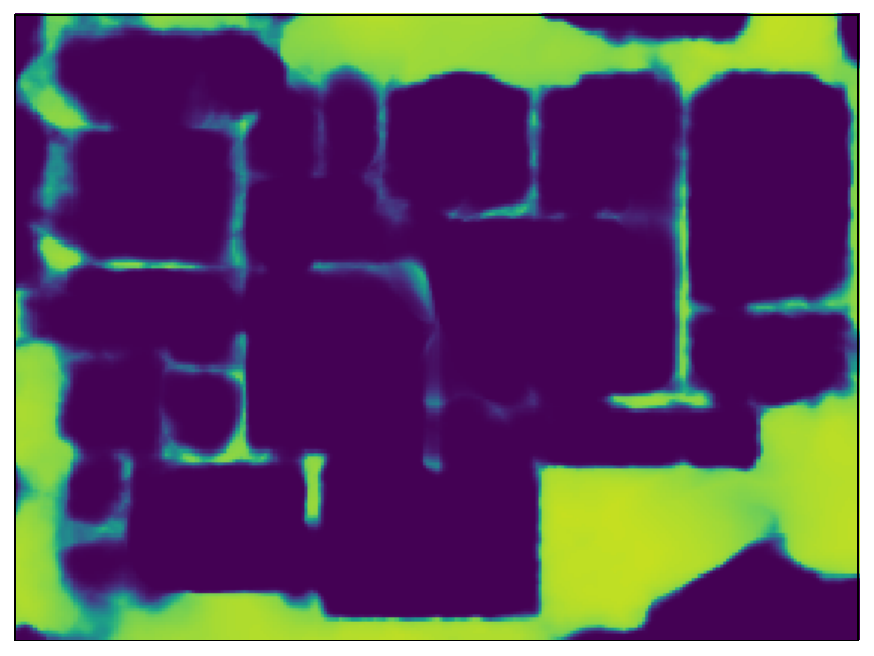}
        \caption{DiNNO, 200 passes}
        \label{subfig:occupancy_mapping:dinno_200_epoch}
    \end{subfigure}
    \caption{{\bf Distributed occupancy mapping.} Our method converges to an accurate map in a single pass through the robots' trajectories, as observations are fully fused into the model. In contrast, DiNNO trains NNs iteratively and therefore requires many passes to overcome catastrophic forgetting. We note that our map at convergence is also more accurate than DiNNO's.}
    \label{fig:occupancy_mapping}
    \vspace{-9pt}
\end{figure*}

We now evaluate on the $2$D online mapping simulation from Yu et al. \cite{yu2022dinno}. The trajectories of $7$ robots around a $2$D environment (Fig.~\ref{subfig:occupancy_mapping:ground_truth}) are predefined\footnote{See Fig. 3 of \cite{yu2022dinno} for trajectories}, and each robot makes a sweep of simulated LIDAR measurements from their position at each timestep. The aim is to use these observations, and inter-robot communication, to learn a map online. To simulate streaming data, Yu et al. \cite{yu2022dinno} give each robot a buffer of the $400$ most recent scans to train its model (total scans per trajectory $\sim4\times10^3$). However, each robot has to repeat its trajectory $\sim100$s of times to achieve a good map. We believe this is due to catastrophic forgetting in NNs, and ask whether our probabilistic representation enables greater efficiency.

\begin{figure}[t]
    \centering
    \includegraphics[width=0.22\textwidth]{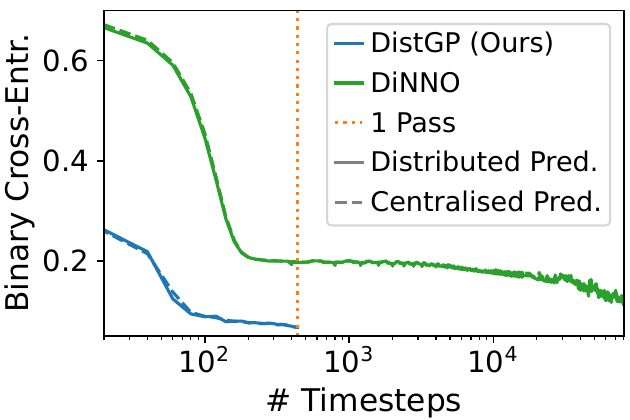}
    \includegraphics[width=0.22\textwidth]{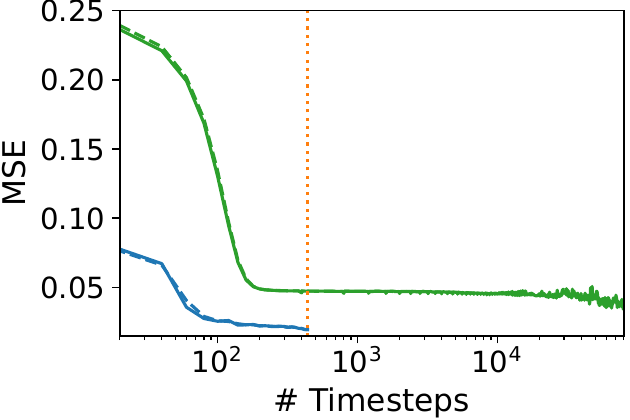}
    \caption{Test error during mapping (log-scale horizontal axis).}
    \label{fig:occupancy_mapping_errors}
    \vspace{-10pt}
\end{figure}
We treat each LIDAR scan of $\sim500$ observations as a batch, use it to update the inducing point posterior, then discard it as per Sec.~\ref{subsec:new_observations} (in contrast to DiNNO's $400$ scan buffer). As the occupancy observations are discrete $y_n\in\{0,1\}$, we use binary observation factors with energy equal to binary cross-entropy (BCE), and $p_n=\sigma(f(\x_n))$ as the predicted probability of occupancy. We linearise these factors to approximate outgoing messages as Gaussian.

Observing that the variation in the map is concentrated at the walls of the environment, we choose inducing points to be close to the walls. From each LIDAR scan we extract a set of candidate inducing points by concatenating: i) the input locations for which density is $1$ (positive examples), and ii) positive examples perturbed a small distance back towards the robot (negative examples). We choose new inducing points from this shortlist using the greedy variance method.

Upon making contact, two robots connect via a shared factor (Sec.~\ref{subsec:async_model_construction}). Each robot also subsamples a set of the other's nearby inducing point locations and creates new inducing points in their own model at these locations. Further, as the trajectories are partially overlapping, the robots exchange inducing point posteriors, allowing them to predict in regions observed by the other without having to incorporate all the other's inducing points into their own model. All robots use a Mat\'{e}rn-$1/2$ kernel with fixed lengthscale.

While our method generates a map distributed between robots, DiNNO produces an estimate of the global map on each robot. We therefore use two modes of evaluation:
\begin{enumerate}
    \item {\bf Global map accuracy}. As in \cite{yu2022dinno}, we assess DiNNO by the average accuracy of all robots' global maps. For our method, we construct the global map which would be achieved by the robots exchanging both i) their own inducing point posterior and ii) posteriors received from others, when connecting.
    \item {\bf Distributed map accuracy.} In each evaluation round, we asses the accuracy when predicting each test point using the closest robot. As the locations change over time, the same test point may be predicted by different robots over the course of the experiment.
\end{enumerate}
The maps (Figs~\ref{subfig:occupancy_mapping:gp},~\subref{subfig:occupancy_mapping:dinno_one_epoch} and~\subref{subfig:occupancy_mapping:dinno_200_epoch}) show that DistGP is able to converge to an accurate map in a single pass, i.e. with each robot traversing their trajectory once. In contrast, DiNNO must repeat the trajectories hundreds of times to get close to the ground truth. We further note that our final map more closely matches the ground truth under both metrics (Fig.~\ref{fig:occupancy_mapping_errors}), and the distributed and centralised predictions perform almost identically for both methods.

\section{Conclusion}
We have presented DistGP: a method for multi-robot mapping in which each robot learns an sparse GP and communicates with others via GBP to ensure consistency. It extends TSGP \cite{bui2014tree} to include loops (which boost accuracy) and support asynchronous GBP. Our experiments demonstrate the inducing point map representation enables effective online data fusion and robust inter-robot communication relative to NN-based DiNNO \cite{yu2022dinno}.

There are many possibilities to build on this work. First, GPs are known to provide accurate uncertainty estimates in many settings, making them effective surrogate models for Bayesian optimisation, and useful for active learning. Our method could be an exciting opportunity for distributed Bayesian optimisation or active learning, in which multiple robots explore different subregions of the input space in parallel. Second, we note that each robot learns kernel hyperparameters in isolation. A message passing procedure to facilitate collaborative estimation in loopy graphs could improve performance considerably.

\section{Acknowledgements}
We are grateful to Javier Yu for his advice on tuning DiNNO.
SN and AJD are funded by EPSRC Prosperity Partnerships (EP/S036636/1) and Dyson Techonology Ltd.

\bibliographystyle{IEEEtran}
\bibliography{IEEEabrv,refs}
\end{document}